# Geoinformatics-Guided Machine Learning for Power Plant Classification

Blessing Austin-Gabriel[1,2], Aparna S. Varde[1,2], Hao Liu[1]

1. School of Computing;

2. Clean Energy and Sustainability Analytics Center

Montclair State University, NJ, USA

(austingabrib1|vardea|liuh)@montclair.edu

ORCID ID: 0000-0002-3170-2510 (Varde), 0000-0002-1975-1272 (Liu)

**Abstract**

While deep learning approaches have advanced power plant analysis from satellite imagery, they are often known to lack the contextual understanding that human experts employ. We thus propose a knowledge-guided machine learning framework leveraging Geographic Information Systems (GIS) with remote sensing in order to enhance the performance of Vision Transformers (ViT) and Convolutional Neural Networks (CNN) in the context of power plant classification. Our main contribution is incorporating spatial binary masks derived from GIS data into the fusion of ViT and CNN architectures to aid power plant classification, demonstrating the potential of geoinformatics-based context for enhancing classification performance. Using real data from large datasets in satellite imagery, we achieve significant classification performance improvements via spatial mask integration, hence paving the way for developing a full-fledged GIS-based knowledge-guided framework to analyze power plants and thus aid future decision-making, mainly in energy management. This work bridges geographic domain expertise with modern deep learning architectures, offering a solid foundation for novel, interpretable, robust environmental monitoring systems.

## Introduction

The integration of domain expertise into AI systems has emerged as a crucial approach in order to improve model interpretability and performance (Varde, 2022). While deep learning has transformed image classification, power plant analysis from satellite imagery presents many challenges that require both visual analysis and geoinformatics-based contexts. Existing approaches often overlook the domain-specific geographic knowledge that human experts subtly use in studying power plant types, locations, and nuances.

Knowledge-guided machine learning (KGML) has demonstrated the potential of incorporating crucial domain knowledge in data-driven AI-based approaches. In recent work, notable researchers (Karpatne et al. 2017) mentioned this paradigm, emphasizing the systematic use of scientific knowledge to enhance AI models. Mainly building upon this foundation, we hereby propose a multimodal framework that integrates GIS (geographic information systems) with deep learning, in the context of power plant classification.

The framework we propose in this paper entails a fusion of Vision Transformers (ViT) and Convolutional Neural Networks (CNN), leveraging both their complementary strengths, and additionally infusing knowledge from GIS in order to harness domain expertise, and enhance power plant classification. This helps to support decision-making in the area of energy management initiatives for sustainability.

CNNs are well-known at extracting hierarchical spatial features (Krizhevsky et al., 2012), while ViTs are known to be effective at modeling long-range dependencies by using attention mechanisms (Dosovitskiy et al., 2021). Thus, in this paper, we leverage the fusion of these two architectures to classify five different types of power plant (wind, solar, natural gas, bio-mass/coal, and hydroelectric) with over 2,262 power plant images from (NAIP, 2017) and (Landsat, 2017) in the (Duke Satellite Dataset, 2017). We compare the performance of a baseline (CNN and ViT only model) with our proposed model incorporating *GIS-derived spatial masks (SM)*, to investigate the feasibility of geoinformatics-guided classification for power plants. Hence, this work is a significant contribution to the main theme of KGML.

Figure 1 presents a partial snapshot of the various types of data analyzed in our work. These include oil, gas, hydro and solar plants, among others.

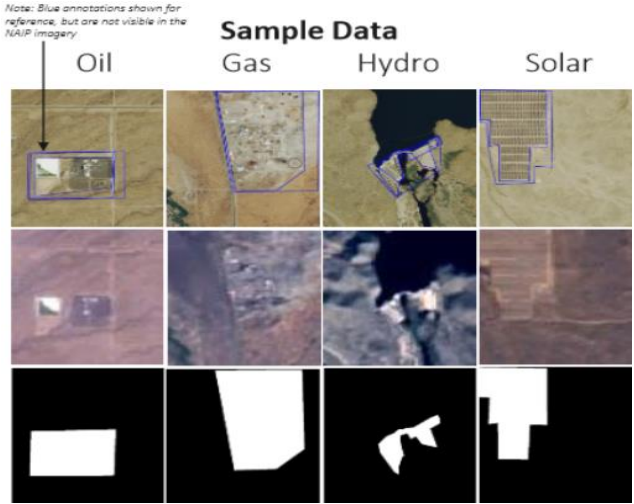

*Figure 1. Sample Data from Different Types of Power Plants*

The main contributions of this research are as follows:
- Proposal of a novel framework adding GIS with domain knowledge to ViT and CNN architectures in order to enhance machine learning.
- Demonstration of spatial mask (SM) integration for improved power plant classification.

## Related Work

There has been much work in the area of domain knowledge adaptation in machine learning studies. For instance, deep learning in remote sensing along with its multiple challenges is discussed (Bell et al. 2017). A systematic review of many web applications in numerous projects on energy-related work is presented in a recent study (Shrestha et al. 2023); while a good review of using domain-specific information from hydrology appears an earlier study (Pathak et al. 2020). Some research (Lin et al. 2023) specifically address CNN-related fault diagnosis of nuclear power plants. Other good research (Zhang et al. 2022) uses ViT mainly to detect fire in power plants. There is interesting work (Shorabeh et al. 2022) using GIS to pick the sites of energy plants.

A few more pertinent studies include: opinion mining in wind energy (Bittencourt et al. 2023); domain knowledge in the diagnosis of turbines (Hu et al. 2016); knowledge-guided deep learning in electric vehicles (Li et al, 2024); use of commonsense knowledge to enhance collaborative tasks among humans and robots in vehicle manufacturing (Conti et al. 2020); greening of data centers with cloud computing for energy efficiency (Pawlish et al. 2015); and optimization of energy systems guided by domain knowledge inclusion (Mahbub et al. 2016).

To the best of our knowledge, our work in this paper is the first of its kind, proposing a novel integration of GIS, CNN, and ViT to enhance power plant classification, aiming to help in better energy management. This reflects the main originality of our work.

## Proposed KGML Framework

Our KGML framework infuses geographic knowledge into deep learning models, leveraging GIS-derived binary masks for ViT and CNN. It is illustrated in a nutshell in Figure 2.

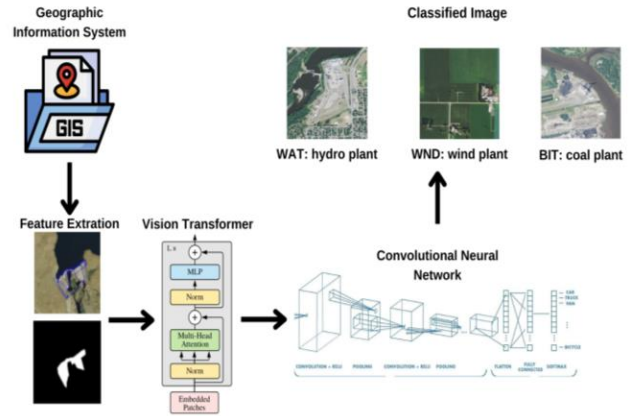

*Figure 2. Proposed Model Architecture: KGML with CNN, ViT and GIS in Power Plant Classification*

Based on this model architecture, our proposed KGML framework is explained briefly in the following parts.

### Local Feature Extraction

The CNN extracts spatial features from binary masks using the given architecture as follows:

$Input \rightarrow Conv2D(64, 3 \times 3) \rightarrow ReLU \rightarrow MaxPool(2 \times 2) \rightarrow Conv2D(128, 3 \times 3) \rightarrow ReLU \rightarrow AdaptiveAvgPool \rightarrow Output \in \mathbb{R}^{\wedge}(14 \times 14 \times 128)$ (1)

This design ensures that CNN output dimensions align with ViT patch embeddings, facilitating effective feature fusion.

### Feature Fusion Strategy

ViT features and CNN features are concatenated as follows:

$Z = [h_{vit}; h_{cnn}] \in \mathbb{R}^{D_v + D_c}$ (2)

The fused representation undergoes dimension reduction via a fully connected layer prior to classification.

## Dataset Curation and Preprocessing

We utilize curated datasets from the USDS: United States Department of Agriculture, NAIP: National Agriculture Imagery Program (NAIP, 2017) and from the USGS: United States Geological Survey, NLCD: New Annual National Land Cover Database (Landsat, 2017) archives. The image datasets mentioned here contain the following:

*Total Images*: 2,262 power plant images

*Categories*: Wind (296), Solar (707), Natural Gas (765), Biomass/Coal (118), Hydroelectric (376)

*Preprocessing*: Standardization (0 mean, unit variance), augmentation (random flips, ±10° rotation), and resizing (224×224 pixels)

## Experiments and Results

We evaluate our proposed KGML framework through a big series of experiments, summarized here. The performance of the original *(CNN+ViT)* model, which is is used as a baseline, is synopsized in Table 1. Notable issues as seen in this model are: ambiguity between visually similar plant types, e.g. bitumen (BIT) versus natural gas (NG); fairly limited performance on classes needing domain context, e.g. solar (SUN) vs. hydro (WAT); and difficulty in wind (WND) farm identification as compared to other types.

With the KGML model entailing GIS-based knowledge integration, the enhanced CNN and ViT with Spatial Masks *(CNN+ViT+SM)* achieve much improvement over the baseline. For instance, some improvement can be observed in wind farm detection. Also, this model offers much better distinction among very similar facility types, e.g. BIT versus NG. The KGML model results are synopsized in Table 2.

*Table 1: Performance Metrics with Baseline Only (CNN+ViT)*

| Category | Precision | Recall | F1-Score | Support |
|---|---|---|---|---|
| WND | 0.89 | 0.81 | 0.85 | 89 |
| SUN | 0.73 | 0.88 | 0.81 | 212 |
| BIT | 0.61 | 0.14 | 0.25 | 35 |
| NG | 0.75 | 0.79 | 0.77 | 230 |
| WAT | 0.75 | 0.81 | 0.78 | 113 |

*Table 2: Performance Metrics with our proposed KGML Model (CNN+ViT+SM)*

| Category | Precision | Recall | F1-Score | Support |
|---|---|---|---|---|
| WND | 0.94 | 0.85 | 0.89 | 89 |
| SUN | 0.87 | 0.92 | 0.88 | 212 |
| BIT | 0.81 | 0.40 | 0.48 | 35 |
| NG | 0.80 | 0.82 | 0.81 | 230 |
| WAT | 0.82 | 0.91 | 0.87 | 113 |

As clearly observed here, our proposed KGML model with CNN+ViT+SM is better in performance across all the power plant types studied in this work. Hence, this work highlights the significant role of the geoinformatics-guided machine learning paradigm.

Motivated by the success of the exploratory research in this paper with our proposed KGML model, we anticipate that there is potential for even further enhancement, which we can address in future research. For instance, the concept of transfer learning has been very successful in some of our earlier work (Karthikeyan et al. 2020) with domain-specific datasets. Further work on these lines might potentially be applicable here. Moreover, we can explore other GIS-based models as a part of our ongoing research.

## Conclusions and Future Work

Our study in this paper demonstrates the scope of integrating vital geographic knowledge into deep learning models for power plant classification to enhance decision support on energy management in the broad context of sustainability. More specifically, the inclusion of spatial masks improves performance, hence validating the crucial importance of geoinformatics in the machine learning process.

As future work, we plan to incorporate real-time GIS data (e.g. wind speeds, solar irradiance) to further improve the model performance. Moreover, we also plan to compare the serial, parallel, and ensemble arrangements of paradigms (GIS, CNN, ViT) for potential enhancement. Our work in this paper is exemplary towards KGML. It has a knowledge-guided method via GIS and domain expertise, and machine learning through a combination of CNN and ViT models. It is anticipated that our exploratory study in this paper can pave the way for much more work along these lines, hence further emphasizing the importance of knowledge-guided machine learning, i.e. the KGML paradigm.

## Acknowledgments

Dr. Aparna Varde acknowledges her NSF grant 2018575 MRI: Acquisition of a High-Performance GPU Cluster for Research and Education. Dr. Hao Liu acknowledges his startup funds from Montclair State University, NJ.